\documentclass{article}

\usepackage[preprint]{neurips_2023}

\usepackage[utf8]{inputenc} 
\usepackage[T1]{fontenc}    
\usepackage{hyperref}       
\usepackage{url}            
\usepackage{booktabs}       
\usepackage{amsfonts}       
\usepackage{nicefrac}       
\usepackage{microtype}      
\usepackage{xcolor}         
\usepackage{graphicx}
\usepackage{enumitem}
\usepackage{multicol}
\usepackage{multirow}
\usepackage{diagbox}

\usepackage{color}
\definecolor{gray}{rgb}{0.5,0.5,0.5} 

\title{WALL-E: \textcolor{gray}{E}mbodied Robotic \textcolor{gray}{WA}iter Load Lifting with \textcolor{gray}{L}arge \textcolor{gray}{L}anguage Model}

\author{%
  Tianyu Wang \\
  Fudan University \\  
  \And
  Yifan Li \\
  Fudan University \\
  \And
  Haitao Lin \\
  Fudan University \\
  \And
  Xiangyang Xue \\
  Fudan University \\
  \And
  Yanwei Fu\thanks{Corresponding author: yanweifu@fudan.edu.cn} \\
  Fudan University \\
}

\begin{document}

\maketitle


\begin{abstract}

    Enabling robots to understand language instructions and react accordingly to visual perception has been a long-standing goal in the robotics research community. Achieving this goal requires cutting-edge advances in natural language processing, computer vision, and robotics engineering.
    Thus, this paper mainly investigates the potential of integrating the most recent Large Language Models (LLMs) and existing visual grounding and robotic grasping system to enhance the effectiveness of the human-robot interaction. We introduce the WALL-E (Embodied Robotic WAiter load lifting with Large Language model) as an example of this integration. The system utilizes the LLM of ChatGPT to summarize the preference object of the users as a target instruction via the multi-round interactive dialogue. The target instruction is then forwarded to a visual grounding system for object pose and size estimation, following which the robot grasps the object accordingly. We deploy this LLM-empowered system on the physical robot to provide a more user-friendly interface for the instruction-guided grasping task. The further experimental results on various real-world scenarios demonstrated the feasibility and efficacy of our proposed framework. See the project website at: \href{https://star-uu-wang.github.io/WALL-E/}{star-uu-wang.github.io/WALL-E/}
    
\end{abstract}

\section{Introduction}

Robots' ability to understand and respond to human instructions is vital, particularly in service tasks. The remarkable progress in Large Language Models (LLMs) has endowed them with the exceptional capacity to comprehend and generate natural human instructions. This advancement should empower robots to learn user preferences and deliver service that closely resembles human interaction. However, despite recent strides in LLMs, there are still challenges that need to be overcome in order to fully realize this potential.


Our work makes significant contributions to addressing the challenges of instructing robotic motion through language, categorizing them into four key aspects.
Firstly, we tackle the issue of \textit{conversational ability}, where our model not only comprehends and generates natural human language but also possesses contextual memory capabilities to retain earlier prompts and prevent information loss. This allows for more coherent and effective dialog between humans and robots.
Secondly, we address the crucial \textit{visual grounding ability}, which involves establishing a meaningful connection between linguistic information and relevant visual context. By bridging the semantic meanings of outputs from LLMs with the requirements of robotic systems in real-world scenarios, we ensure a better alignment between language instructions and visual contexts.
Thirdly, we focus on the \textit{primitive action ability} of robots. This involves enabling the robot to execute a sequence of fundamental actions, such as grasping, screwing, and placing objects, based on preliminary visual understanding. By enhancing the robot's capability to interact effectively with the real environment, we enable it to perform complex tasks with greater efficiency and accuracy.
Lastly, we resolve the \textit{language-motion alignment ability} of robot. This bridges the wide gap between the high-level planning of LLMs and the low-level execution of real-world robotic system.
Such alignment is achieved by utilizing the vision-and-language models to ground target object for robotic grasp, transforming the output of LLM into actual grasp target.

While previous works~\cite{cheang2022learning, shridhar2022cliport, shridhar2023perceiver} have made strides in addressing these challenges, they have predominantly relied on simple and concise referring expressions. Our work goes beyond these limitations by tackling the complexities of lengthy human-robot dialogues and incorporating feedback on the state of the environment after robotic operations. This unlocks the potential for robots to handle more intricate tasks and enhances their overall performance and adaptability.

\begin{figure}
    \centering
    \includegraphics[width=0.9\columnwidth]{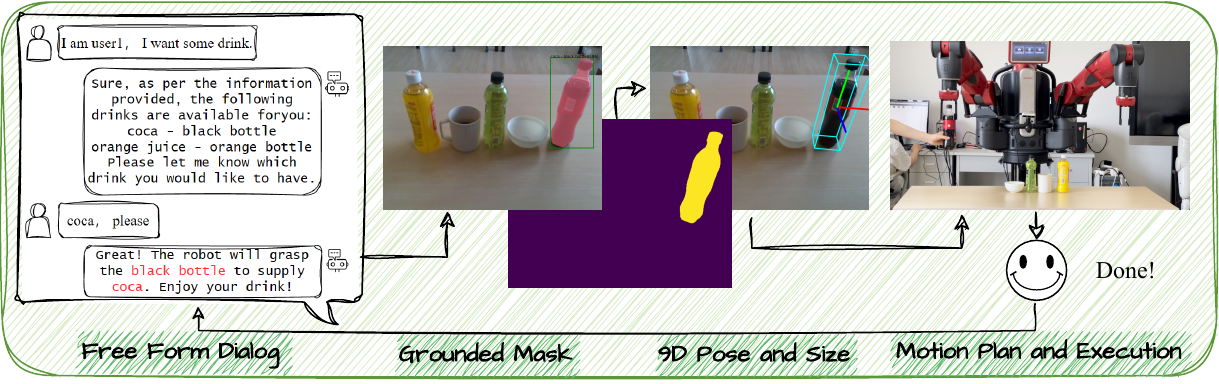}
            \vspace{-0.1in}
    \caption{Task description of our framework. The framework first gives the target instruction that indicates the preference of user via the multi-round dialogues. Then the target object mask is grounded according to the target instruction, after which the detailed 6-DoF object pose and 3D size are predicted for calculating the gripper pose. Finally, the robot will plan and execute feasible motion to the target pose for grasping specific object.  \label{fig:teaser}}
        \vspace{-0.15in}
\end{figure}

Formally, in this paper we introduce \textit{WALL-E}: \textbf{E}mbodied robotic \textbf{WA}iter load lifting with \textbf{L}arge \textbf{L}anguage model, which makes a significant contribution to the enhancement of usability in robotic systems by harnessing the power of LLMs. Our main focus is on creating a more user-friendly human-robot interface, and to achieve this, we propose a novel framework that integrates LLMs into robotics research. Specifically, 
our approach explores the combination of off-the-shelf visual grounding and robotic systems, leveraging the capabilities of the LLM ChatGPT. This integration surpasses the performance of previous independent systems in terms of instruction understanding, leading to more effective human-robot interaction.

To initiate the process, we guide ChatGPT with a prompt and constrain its role to a specific domain. We formalize the operating environment in a dictionary format and provide it as input to ChatGPT. Experimenters then engage in interactive dialogues with ChatGPT over multiple rounds. Through this extensive interaction, ChatGPT synthesizes the experimenters' preference object into a target instruction. The summarization is carefully validated by the experimenters before being passed on to the next robotic system.

The resulting target instruction, as illustrated in Fig~\ref{fig:teaser}, is then fed into a 9D visual grounding sub-system of extracting the 6 Degree-of-Freedom (DoF) and 3 Dimension (D) size of the target object. 
This system consists of several key components in vision community such as DINO~\cite{liu2023grounding}, SAM~\cite{kirillov2023segany}, and SAR-Net~\cite{lin2022sar}. By grounding the target object and determining the grasp location, this visual grounding system enables the real-world robot to execute the top grasp on the target object, even in the presence of other obstacles.

Subsequently, the experimenter provides feedback on the status of the robotic grasp, allowing ChatGPT to continue its operations based on the memory of the most recent robotic actions. It is crucial to highlight that our proposed system-level framework is a pioneering effort, designed specifically to unlock the potential of robots executing actions guided by long-step instructions. Moreover, with its impressive memory capabilities, this system exhibits exceptional interactive abilities, facilitating seamless dialogue with various numbers of users.

In this paper, we present several noteworthy contributions:

 \textbf{1.}  We introduce a groundbreaking system that combines the conversational ability of ChatGPT with the object grounding and robotic grasping capabilities of off-the-shelf system, realizing the well alignment of LLM's output instruction to actual robotic action. This integration demonstrates the remarkable potential of leveraging LLMs in conjunction with existing robotic systems, paving the way for more advanced and engaging human-robot interactions.  \\
 \textbf{2.}   A novel 9D visual grounding sub-system
   is presented to conduct visual alignment by mapping linguistic information to relevant visual cues, enabling the whole system to understand the meaning of language instructions within a visual context.\\
 \textbf{3.}  We implement and deploy the LLM-empowered grasping system on a physical robot. By doing so, we provide a significantly improved human-robot interactive interface compared to previous systems. This enhancement enhances the user-friendliness and effectiveness of the overall system. \\
 \textbf{4.}   We conduct thorough evaluations of our system in real-world scenarios. These evaluations serve to validate the feasibility and utility of our approach, ensuring that our system performs reliably and effectively in practical applications. 

Collectively, these contributions highlight the innovative nature of our system, its practical deployment on a physical robot, and the successful validation of its performance in real-world settings.

\section{Related Work}
\textbf{Large Language Models}.
Large Language Models (LLMs) are a type of language model that have shown exceptional performance in various natural language processing (NLP) tasks.
Some of the recent LLMs include GPT-3, PaLM, Galactica, LLaMA, GPT-4 and PaLM 2~\cite{chowdhery2022palm, touvron2023llama, taylor2022galactica, bowman2023eight}. These models have been used to generate realistic text with varied topics and sentiments. LLM has strong emergent abilities, including contextual understanding and step-by-step reasoning. For small language models, it is usually difficult to solve complex tasks that involve multiple reasoning steps, including mathematical solutions, understanding user intentions, and other intrinsic hallucination~\cite{zhao2023survey}. In addition, the prompt strategy assisted by chain-of-thought (COT), which involves intermediate reasoning steps for deriving the final answer, can further unleash the reasoning ability of LLMs~\cite{zhang2022automatic, kojima2022large, zhou2022least, press2022measuring, wang2022self}. LLMs and prompt engineering enable us to easily establish high-level human-robot interactions and support a wide range of robotic manipulation tasks by obtaining language goals.

\textbf{Visual Grounding for Robotics}
Visual grounding involves localizing the target object from the image based on the given linguistic description. Recent works~\cite{guadarrama2014open, nguyen2020robot, shridhar2018interactive, yang2021attribute, cheang2022learning, chen2021joint} have shown promising results on retrieving the target object for grasping by using the text instruction as a query. However, these methods are limited in their ability to handle more complex and longer referring expressions and to support interactive feedback from humans.
To address these limitations, recent proposed methods, such as~\cite{zhang2021invigorate, shridhar2020ingress, yang2022interactive}, have developed interactive grasping systems to resolve ambiguities via dialogues. However, these methods have limited capacity for memory and reasoning, relying on partially observable Markov decision processes. In contrast, our proposed framework takes advantage of the LLM, which has powerful long-term memory and reasoning abilities, to improve upon existing methods for human-robot interaction.

\textbf{LLMs for Robotics}
Embodied intelligence mainly focuses on building systems where agents can purposefully exchange energy and information with the physical environment. It requires a correct understanding of the embodied perception process from a high-dimensional cognitive perspective to a low-dimensional execution perspective~\cite{roy2021machine}. Recent work has shown that using LLM as robotic brain can unify egocentric memory and control by studying downstream tasks of active exploration and embodied question answering~\cite{mai2023llm}. However, the perception system included in this new framework still has shortcomings in visual grounding design, which potentially limits the interaction between robots and the environment. On the other hand, there are zero-shot or few-shot methods ~\cite{vemprala2023chatgpt, wake2023chatgpt, brohan2023can, huang2022inner, huang2023grounded, wu2023tidybot} that utilize LLM as task planners, decomposing high-level instructions into executable primitive tasks. These methods assume the ability of robots to execute advanced commands. However, they cannot support open-vocabulary interaction with robot and not robust enough due to insufficient perception of  environment.
Instead, our framework effectively addresses these two issues and provides a flexible paradigm to bridge LLM with agents, suggesting a new dimension of thinking for universal human-robot interaction.

\section{Approach}
\textbf{Task Definition.}
We assume the system has access to robotic operating environment and objects. Thus, we parser this scene in a formalized style, and represent it as a dictionary. Then, we use the LLM of ChatGPT to summarize the user preference as an target instruction.
Given this instruction and an RGB-D image of the scene, our goal is to ground the target object from the linguistic descriptions for robotic grasping. 

\begin{figure}
    \centering
    \includegraphics[width=0.9\columnwidth]{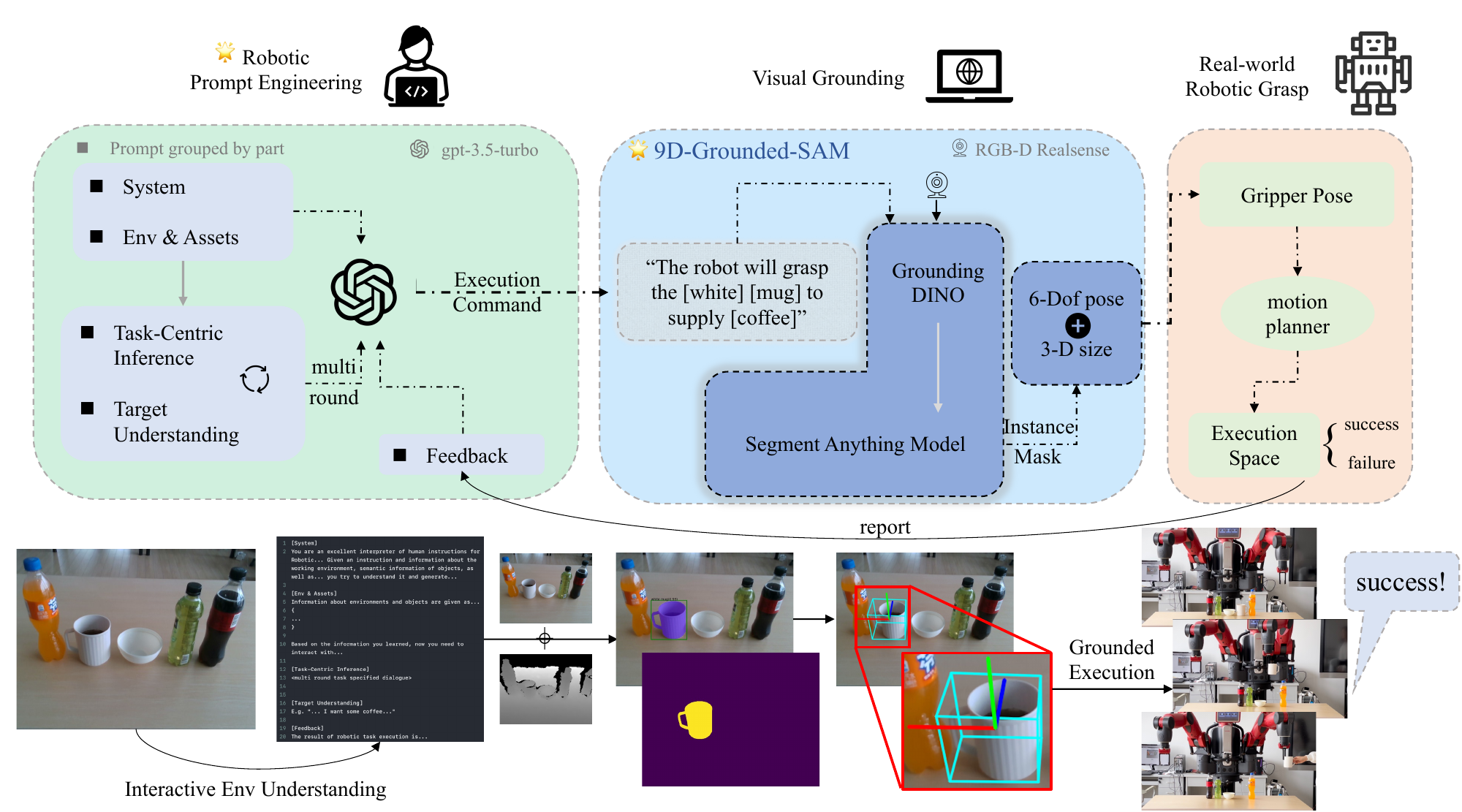}
            \vspace{-0.1in}
    \caption{An overview of our proposed WALL-E system. Given the prompt of role and environment, the ChatGPT first summarizes the preference object of the users as a target instruction through the multi-round dialogue. The instruction is further forwarded to the 9D-Grounded-SAM, where the Grounding DINO and SAM is first utilized to localize the mask and label of target object, and then SAR-Net is adopted for object 6-DoF pose and 3D size estimation. Furthermore, the target robotic gripper pose is calculated, after which the motion planning is executed for object grasp. Finally, the user gives feedback on the state of the environment after robotic operations. \label{fig:pipeline}}
        \vspace{-0.15in}
\end{figure}

\textbf{Overview of the System.}
To learn user preferences of objects from the multiple round of the human-robot interaction of text, our WALL-E system has three main components containing the instruction understanding, object grounding, and robotic grasping as in Fig.~\ref{fig:pipeline}.
We first take the advantages of the recent ChatGPT by given prompt of the task context. Then we develop an user interface where takes the user instruction into the LLM in the form of text. Thus the user can conduct multi-round dialogues according to the textual output of the model. According to such a multi-step human instructions, the model finally captures the user preference of the object on the current robotic task space, e.g.,``The robot will grasp the white mug to supply coffee" as in Fig~\ref{fig:pipeline}.  Then, the object grounding component infers the potential object mask according to the language instruction, and this mask is further sent to the pose estimation modules to recover the object pose and size in the camera coordinate. Finally, the grasp pose of gripper is calculated, and the gripper motion is planned by using the motion planning algorithm to reach the target gripper pose for grasping.

\subsection{Prompt Design for LLMs}
\begin{figure}
    \centering
    \includegraphics[width=1.0\columnwidth]{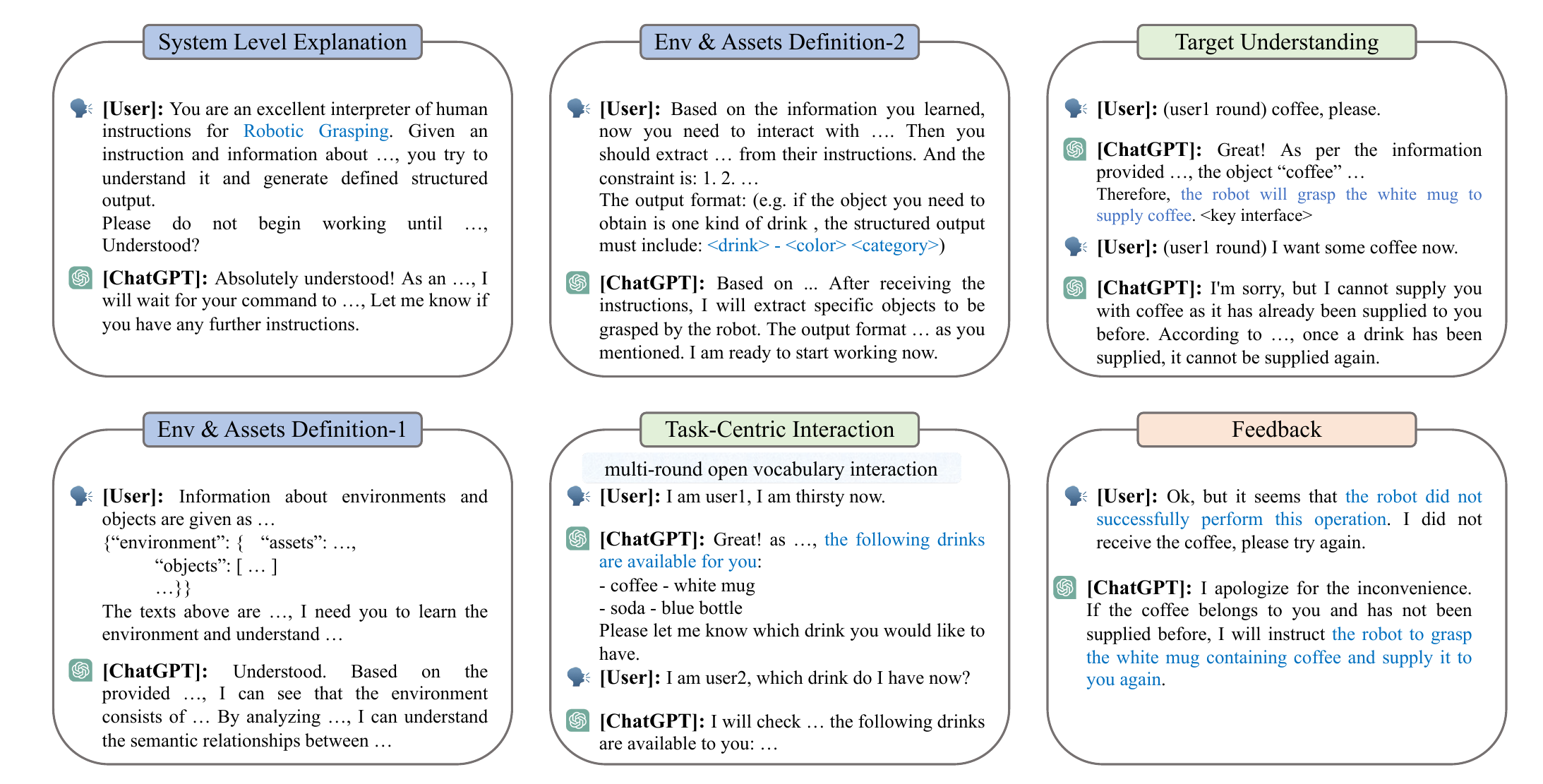}
        \vspace{-0.2in}
    \caption{The design principle of robotic prompt. Hierarchical prompts are presented in different panes, which interact with ChatGPT in a step-by-step manner. We try to present only one modular example in each part of prompts for brevity. \texttt{The system explanation and definition of environment space and task space} (in blue), \texttt{multiple rounds of task-centric interaction and command generation} (in green), and \texttt{feedback mechanism} (in fleshcolor) are example completions. 
    \label{fig:prompt}}
    \vspace{-0.2in}
\end{figure}

Our proposed human-robot interaction mode empowered by LLMs demands a hierarchical prompt design principle which is designable for specific tasks and able to support multiple rounds of open vocabulary dialogue. We take ChatGPT as the default LLMs in this work.
The recent work on prompt design has provided a lot of inspiration~\cite{wake2023chatgpt, zamfirescu2023johnny, wei2022chain}. However, it should be pointed out that these designs often only consider how to limit the LLM to the task space and generate specific language goals, without ensuring the flexibility of human-robot interaction. We are trying to bridge this gap on an open vocabulary domain. Based on the task space of robotic grasping and our desire for flexibility in human-robot interaction, we have designed a set of prompts as in Fig~\ref{fig:prompt}, which consist of (1) the explanation and functional guidance for the role of ChatGPT, (2) the definition of environment and semantic information of assets, and the description of task rules and restrictive output information for ChatGPT, (3) the task-centric open vocabulary communication, (4) the confirmation of target before execution command generation, and (5) the feedback on robotic execution space. In such hierarchical prompt engineering, the (1) and (2) often exist as prior knowledge, equivalent to a process of guiding ChatGPT learning. The point 3 and point 4 are presented in the form of open vocabulary multiple rounds of dialogue, with the point 4 being a key interface in human-robot interaction. The point 5 is the report on results of task execution, which includes corresponding fault tolerance mechanisms. Detailed examples of prompts and dialogues are provided in the supplementary materials.

\noindent\textbf{System Level Explanation.}
In the first part of the prompt, we provide a system level explanation for ChatGPT. This explanation aims to guide ChatGPT to confirm its role and limit it to specific domain, in order to ensure the robustness and professionalism of task analysis in the robotic grasping. Simultaneously, as we hope to carry out multiple rounds of open vocabulary interaction in various robot sub-tasks, a text is also provided for ChatGPT in this part, reminding him that his main task before interacting with the user is to understand and learn the environment and the specific task, rather than directly starting work.

\noindent\textbf{Environment and Assets Definition.}
The second part of the prompt is divided into two sub modules, one supporting the representation of the environment and assets, and the other supporting the interpretation of task rules and limiting conditions and output formats. In the first module, we leveraged a method of interactive environment understanding, which provide a more reliable auxiliary learning. The asset space and object semantic space together constitute the environment space. The environment was represented as a dictionary containing the lists of assets, objects and their visible semantic information as in Fig~\ref{fig:prompt}. Semantic information is to encapsulate the understanding of images in objects, so that ChatGPT can capture each object and provide a knowledge base for subsequent inference. In the second module, we first provide a further description of the task space and prompt for key information that needs to be extracted from user instructions. We provide a segmented description of the constraints of the task, expecting that these rules and constraints will be continuously checked during the interactive dialogue. Finally, we introduced the expected output format. In order to maximize the retention of open vocabulary communication and improve scalability for unprecedented robotic manipulation task interactions, we allow the inference process to be simply expressed. However, the output of  structured key information is still important. In our grasping task, We require ChatGPT to provide key information through forms such as <object> - <color> <category>, which will be transmitted to visual grounding. In addition, this form also provides the possibility for further exploration of the application of chain of thought and synthetic prompt in human-robot systems~\cite{shao2023synthetic}. It should be pointed out that the definitions of environment space and task space in the second part serve as dependencies, forming the foundation of execution space.

\noindent\textbf{Task-Centric Interaction.}
The prompts in the third part are designed to support multiple rounds of open vocabulary interaction. The example given in Fig~\ref{fig:prompt} is related to the multi user management of the task. User input should be as unrestricted as possible, which is the pursuit of human-robot interaction. Users engage in conversations with ChatGPT through daily open vocabulary, and then it relies on task space for inference and dialogue with users. In this part of the prompt, we expect to explore the robustness of the LLM's memory and long-step reasoning abilities towards the environment, tasks, and interactions through multiple rounds of dialogue.

\noindent\textbf{Target Understanding and Command Generation.}
The fourth part of the prompt is a key interface between the large language model and visual grounding. In our task, when the user expresses a direction towards the target, ChatGPT understands the target in the task space and generates structured output, which will serve as an execution command. This command exists in the form of open vocabulary, and When it forms an image-text pair with visual space and then the target output is generated through a grounded model, the accuracy of the target output is a part of our evaluation of the effectiveness of the integration of LLM and the vision model.

\noindent\textbf{Feedback from Execution Space.}
We designed the last part of the prompt to set up a feedback mechanism for human-robot interaction, which enables ChatGPT to timely perceive the task space and update the context by reporting the result observation of the execution space. The feedback prompt means that if the task is successfully executed, the user will provide confirmation feedback to facilitate subsequent rounds of dialogue. If the task fails to execute, the user will report that the execution was not successful, and ChatGPT will need to perform actions such as backing off the context information in the environment space.

\begin{figure}
    \centering
    \includegraphics[width=1.0\columnwidth]{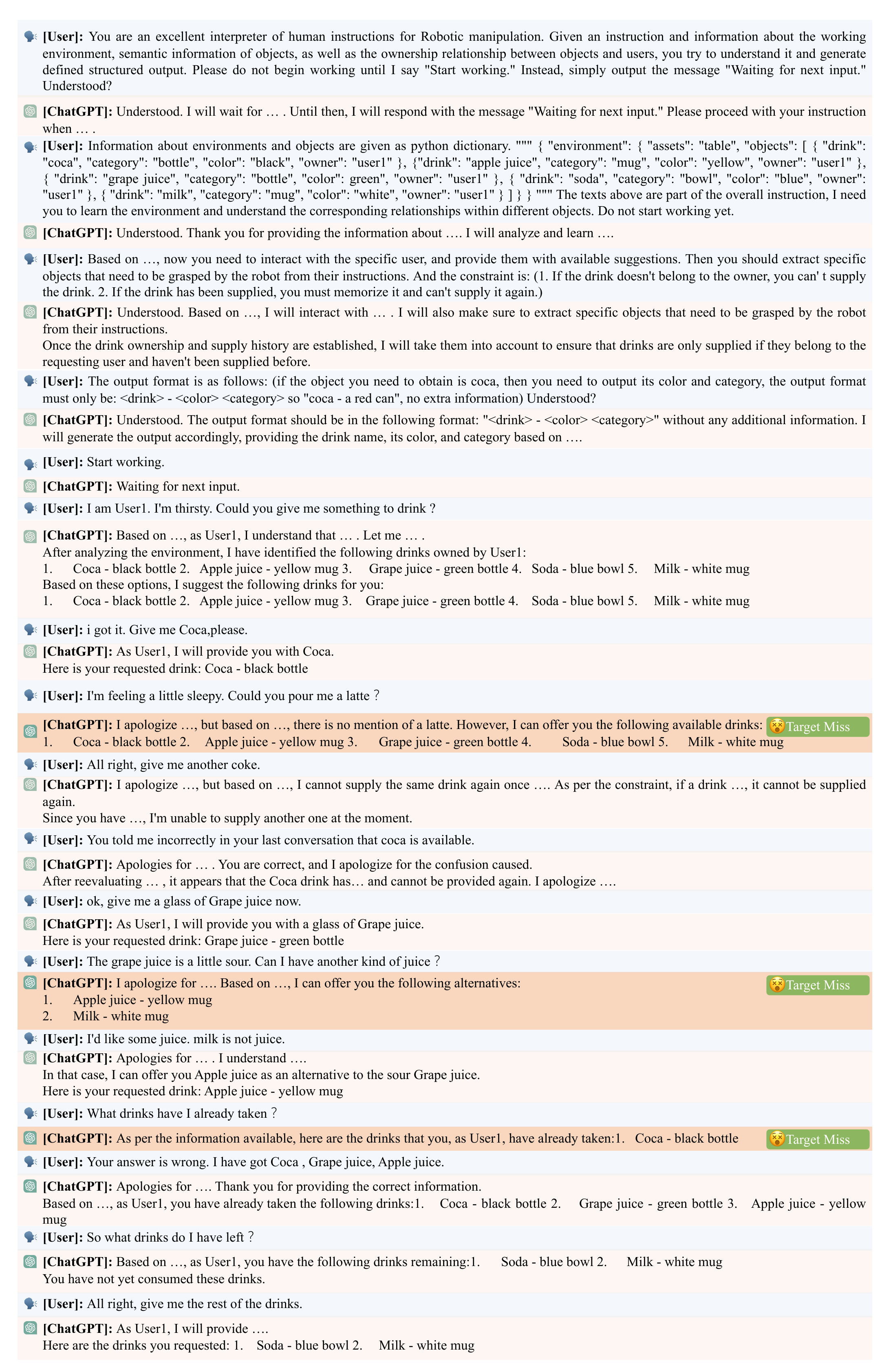}
            \vspace{-0.1in}
    \caption{The figure shows the scenario when there is only one user. Some of the redundant information we have omitted in the presentation diagram. Open-vocabulary linguistic interaction module may make errors in the inventory of environmental items, in the suggestion of user's needs, and in the duplication of items due to memory confusion.\label{fig:demo user1}
        \vspace{-0.15in}}
\end{figure}

The following case in Fig.~\ref{fig:demo user1} shows how WALL-E handles one user's requests in a scenario where there are five drinks including coca, apple juice, grape juice, soda and milk on the table. This dialog demonstrates the great freedom and strong robustness in the process of communication. Although ChatGPT sometimes makes mistakes, with the feedback from user, it can quickly correct the confusion and get the final expecting robotic commands.

\subsection{9D-Grounded SAM for Visual Grounding}
\noindent\textbf{2D Visual Grounding.}
When users engage in dialogue with ChatGPT, they type their instruction onto the terminal, and the language model produces a target instruction that identifies the preferred object. This instruction is then sent into Grounding DINO~\cite{liu2023grounding}, an open-set object detector that uses vision-language modality fusion to generate bounding boxes and labels with free-form referring expressions. This process involves multiple phases, including a feature enhancer, a language-guided query selection module, and a cross-modality decoder. Once the grounding box of the target object is obtained, it is used as a prompt for the Segmented Anything Model (SAM)~\cite{kirillov2023segany}, a powerful object segmentation model that generates refined object masks. This mask is then used to crop the depth pixels belonging to the object, facilitating follow up pose and size estimation.

\noindent\textbf{6-DoF Object Pose and 3D Size Estimation.}
After obtaining the grounding mask and category label of the target object, we use a pose and size estimation module to recover the object 6-DoF pose and 3D size in the camera coordinate system, which provides information about the object's state in the current task space. The object pose consists of a 3-DoF translation and a 3-DoF rotation, while the size refers to the dimensions of the object's bounding box in the real world. Recent advances in object pose estimation have enabled category-level generalization across instances within a category, making it intuitive to utilize off-the-shelf category-level object pose estimation models. We have chosen to use SAR-Net~\cite{lin2022sar}, a light-weighted model that runs in real-time. By providing the depth image, mask, and category label of the target object to SAR-Net, we can obtain the object 6-DoF pose and 3D size. As the camera is eye-hand calibrated, the estimated pose can be easily transformed into the robot's base coordinate system for further motion planning.

\textit{Why is it necessary to consider the object's pose instead of using a straightforward grasp pose estimation method?}
While recent methods for straightforward 6-DoF grasp pose estimation~\cite{mousavian20196, fang2020graspnet, fang2022anygrasp} have made significant progress, they remain limited in scope and are mainly suitable for pick-and-place operations when deployed on real robots. In our system, we integrate object pose estimation methods, as the object's state in 3D space is critical for calculating meaningful operating points in various manipulation tasks, such as door opening, water pouring, and hole pegging. We believe that object pose provides rich context for robots before motion planning in manipulation tasks. To ensure the extensibility of our system, we continue to follow the principle of inferring the object pose first and then calculating the useful target gripper pose for elaborate manipulation.

\subsection{Robotic Grasping}
\begin{figure}
    \centering
    \includegraphics[width=0.9\columnwidth]{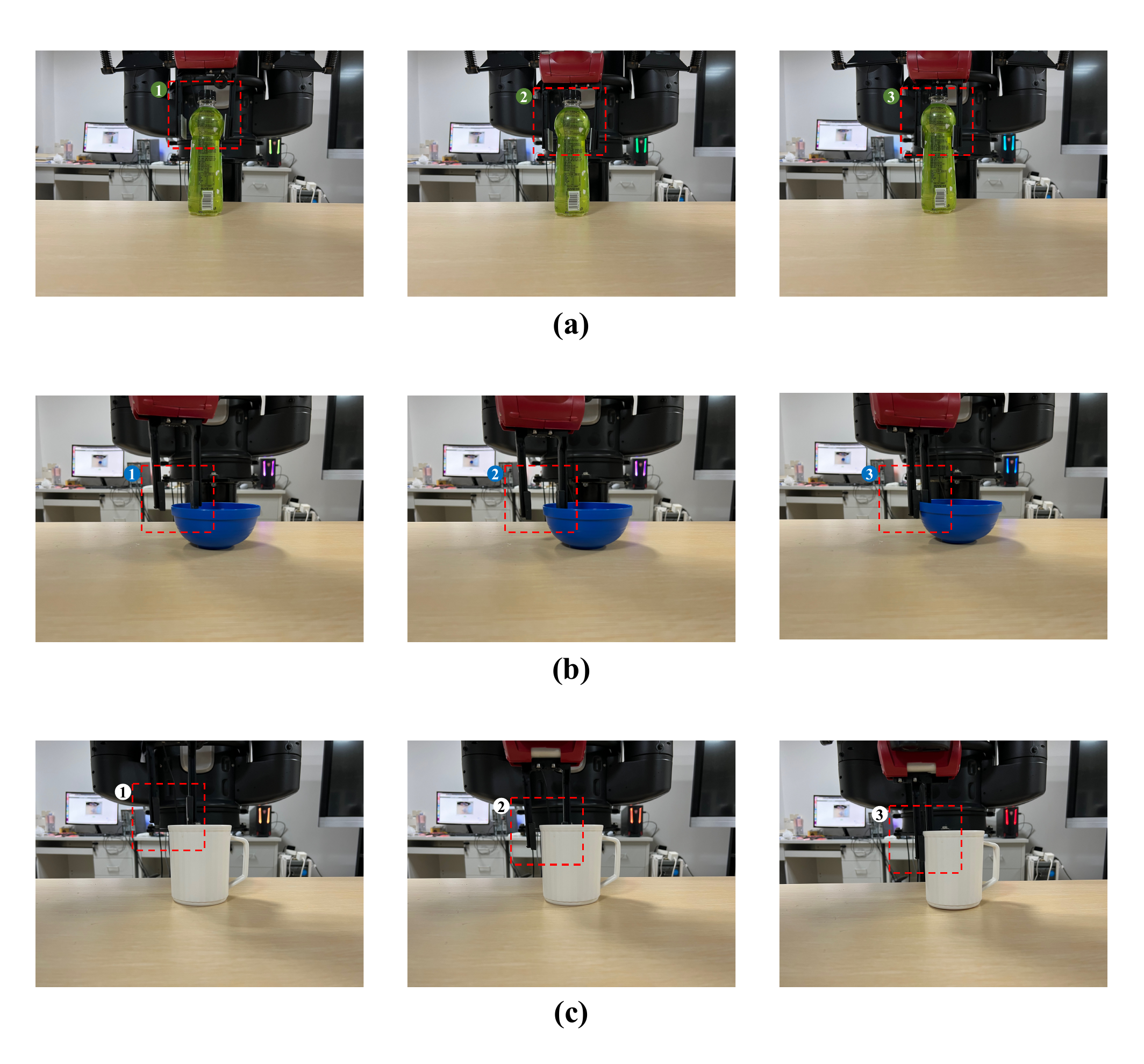}
            \vspace{-0.15in}
    \caption{\textbf{Examples of the top-grasp policy on different category of object.} Examples of different grasp details in bottle (a), bowl (b) and mug (c). (a): Due to the generally small caliber of bottles, we choose the method of grasping the bottle body based on the 6D pose and 3D size. The grasping position at this point is defined as one-third of the height, and the center of the gripper is aligned with the center of the bottle body. (b) \& (c): For bowls and mugs, since their calibers are relatively large, we choose the method of grasping edges based on point offset.\label{fig:grasp_policy}}
        \vspace{-0.15in}
\end{figure}

We pre-define grasp point on the object of each category, i.e., grasping the center of bottle, and grasp the side body of the bowl and mug. We use the top-grasp policy, thus the orientation of the grasp pose is aligned to the direction of gravity. Considering the different caliber of the containers, including bottle, bowl and mug, we have designed a category-level grasping strategy that can meet a wide range of scenarios. The different details of grasping are mainly reflected in the coordinates of the pick point as in Fig.~\ref{fig:grasp_policy}. Overall, the top-grasp policy determines the grasp points in the robot coordinate system, which vary depending on the category and specific size. Our grasping strategy provides a set of processes for embodied entities, including sub tasks such as picking, lifting, moving to user, and controlling gripper status. In addition, the  top-grasp policy indicates that, during the grasping process, the rotation of the gripper remains unchanged, which is verified to ensure stability in human-robot interaction. Once we obtain the calculated grasp points in the robot coordinate system, we use the motion planning library MoveIt!~\cite{chitta2012moveit} to generate a trajectory from the current robotic gripper pose to the target grasp pose. Finally, MoveIt! calculates the feasible joint motion for the robot to reach the target grasp pose.
\vspace{-0.1in}

\section{Experiment}
\subsection{Setting}
\textbf{Experimental Setup and Details.} Our system is deployed on a Baxter robot with dual 7-DoF arms and parallel grippers. The RealSense D435 camera, mounted on the robot's torso, captures RGB-D streams for the scene. The system operates on a Desktop with an NVIDIA GTX1070 GPU, and we adopt the pre-trained Grounding DINO, SAM and SAR-Net models in our system. 
\noindent\textbf{Principle of Human-robot Dialogue.} 
In the real robot experiment, multiple rounds of dialogue will take place between the experimenters and the system. Each trial involves up to three different experimenters interacting with the system individually. The system then summarizes each experimenter's preference as the target instruction in the format of <object> - <color> - <category>, which is then forwarded to the visual grounding system.

\begin{figure}
    \centering
    \includegraphics[width=0.9\columnwidth]{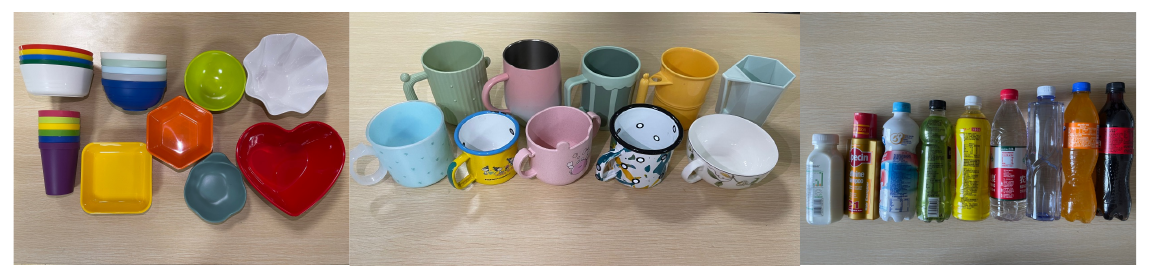}
            \vspace{-0.1in}
    \caption{The 41 objects used in our robotic experiment. We collect real-world objects from three categories including bottle, bowl and mug. These objects within a category show large diversity in textures, calibers, sizes and shapes. \label{fig:objects}}
            \vspace{-0.1in}
\end{figure}

\noindent\textbf{Test Object Collections.} 
Using this robot system, we conducted tests on 41 real-world scenarios, employing a collection of bottles, bowls, and mugs (Fig~\ref{fig:objects}). In each scenario, we randomly selected 5 instances from the object set and placed them on reachable locations on the table in front of the robot. None of the testing objects were utilized for training the models in our system.


\noindent\textbf{Evaluation Metric.} 
To quantitatively evaluate the performance of our system, we propose three metrics: (1) Success rate of instruction targeting. Our LLM-empowered system outputs the target instructions which determine the object that the user wants. We consider a final output instruction a success if it identifies the target object, otherwise a failure. (2) Success rate of visual grounding. After obtaining the target instruction, the visual grounding component localizes the mask of target object. We consider a grounding result successful if the predicted mask visually overlaps with the target object and a failure if it overlaps with non-target object or only has a small overlap ratio with the target object. (3) Success rate of grasping. The robot executes grasp actions based on the calculated target gripper pose. We consider a grasp attempt successful if the gripper grasps the target descriptive object.
Overall, we calculate the success rate as the ratio of successful attempts to the total number of attempts. In our actual real robot experiment, we test our system on object from three categories, and we try 15 attempts on each category.

\begin{table}[]
\centering
\caption{Results of our framework in real robot experiments. We show the success rate tested on object of different category under the three metrics. ``Ins'' means the instruction targeting, ``Vis'' indicates the 2D visual grounding and ``Grasp'' is the robotic grasp, respectively.}
\label{tab:1}
\resizebox{\columnwidth}{!}
{
\begin{tabular}{c|ccc|ccc|ccc|c}
\toprule[1pt]
\multirow{2}{*}{\diagbox[]{User Num.}{Success Rate}} & \multicolumn{3}{c|}{Bowl} & \multicolumn{3}{c|}{Bottle} & \multicolumn{3}{c|}{Mug} & Total \\ \cmidrule{2-10}
 & Ins & Vis & Grasp & Ins & Vis & Grasp & Ins & Vis & Grasp & Grasp \\ \midrule[0.5pt]
1 & 93.33 & 86.67 & 86.67 & 100 & 93.33 & 86.67 & 100 & 73.33 & 60.00 & 77.78 \\ 
2 & 86.67 & 73.33 & 73.33 & 93.33 & 80.00 & 66.67 & 93.33 & 80.00 & 46.67 & 62.22 \\ 
3 & 73.33 & 66.67 & 66.67 & 80.00 & 73.33 & 60.00 & 73.33 & 53.33 & 26.67 & 51.11 \\ \midrule[0.5pt]
Total & 84.44 & 75.56 & 75.56 & 91.11 & 82.22 & 71.11 & 88.89 & 68.89 & 44.45 & 63.71 \\
\bottomrule[1pt]
\end{tabular}
}
\end{table}

\begin{figure}
    \centering
    \includegraphics[width=1.0\columnwidth]{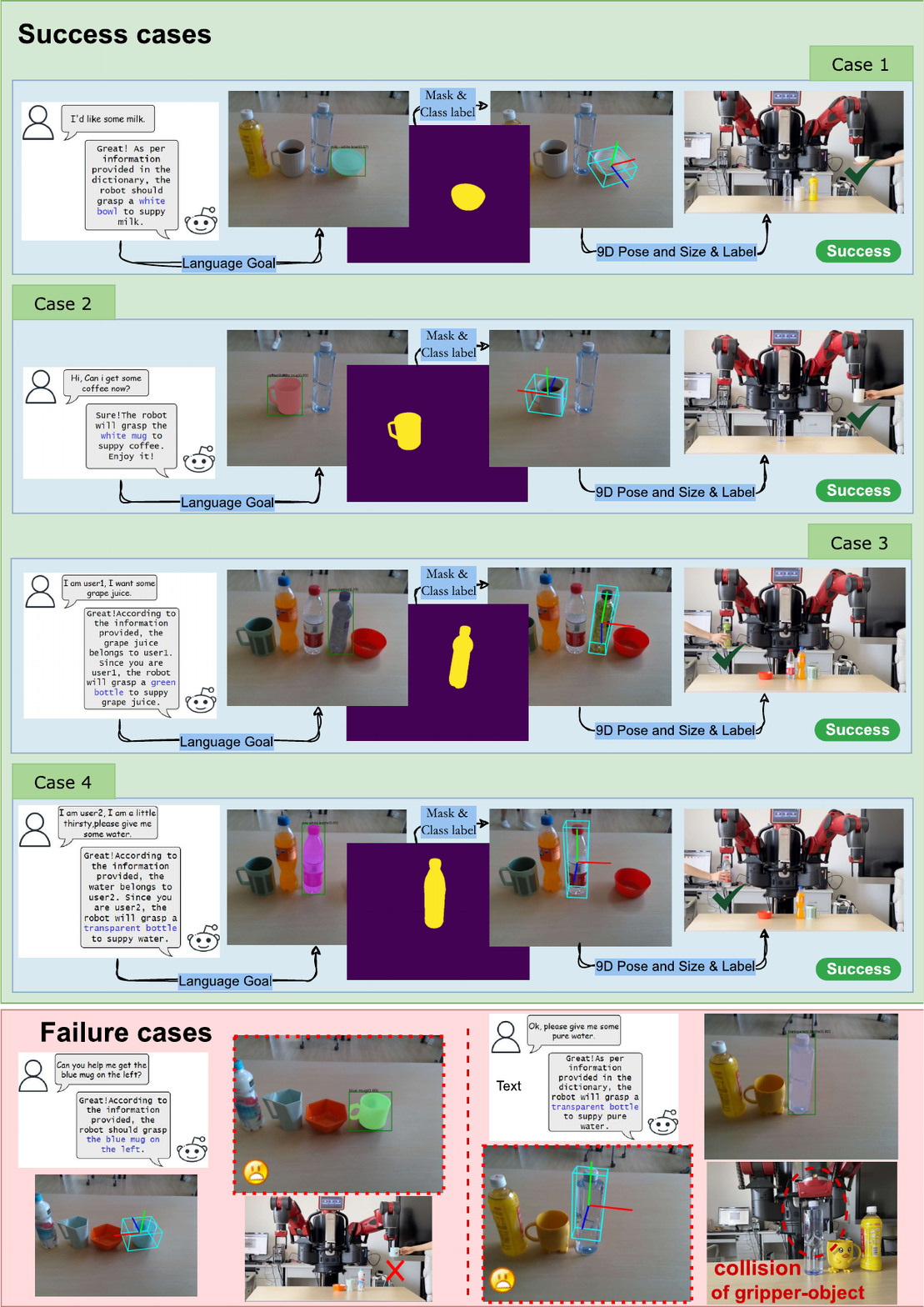}
            \vspace{-0.1in}
    \caption{Visualization of the intermediate results from our systems during the real robot evaluation. The top four rows are success cases of robotic grasp guided by instruction. From the left to right, we show the target instruction from ChatGPT, grounding object mask, oriented 3D bounding box, and state of robotic grasp, respectively. The last row shows the failure cases due to the incorrect grounding mask, and inaccurate pose and size estimation. \label{fig:qualitative results}}
            \vspace{-0.15in}
\end{figure}

\subsection{Experimental Results}
\noindent\textbf{Quantitative Results.} We present a summary of the success rates of the sub-modules of our system in Tab.~\ref{tab:1}. Our results indicate that the LLM exhibits promising performance in multi-round dialogues with single users, with a success rate exceeding 90\%. For multiple users, the results demonstrate the significant memory capacity of the LLM in handling instructions from multiple users. However, as the number of users increases, the success rate of instruction targeting decreases due to the limited scope of memory capacity. However, we observed that the success rate of the final grasp is relatively lower than that of instruction targeting, mainly due to two reasons. Firstly, gripper-object collision can occur despite accurate 9D grounding results of the object, particularly when the target object is located in close proximity to other objects. This collision can impede the successful grasp of the target object. Secondly, infeasible inverse kinematics can occur when the planning algorithm struggles to calculate feasible joint motion for the target grasp pose..

\noindent\textbf{Qualitative Results.} Figure~\ref{fig:qualitative results} showcases the qualitative outcomes of our system in the context of real robot experiments. Our system proficiently summarizes user preferences through input of free form instructions. The further grounding module accurately generates pixel-wise masks of the user-depicted objects and predicts the tightly oriented bounding boxes of the target objects. Our framework benefits from the proposed 9D-grounding-SAM, which enables precise inference of object states even in robotic workspaces that have random arrangements of instances from different categories. However, our system sometimes fails due to inaccurate visual grounding results, as exemplified by the last row of Fig.~\ref{fig:qualitative results}, where the 2D visual grounding component provides incorrect estimations leading to mistaken object grasping. Additionally, the materials of the object may also affect the quality of depth image, resulting in inaccurate object pose and size estimation due to missing depth. As a consequence, the object center may be positioned above the actual one, leading to collisions between the gripper and object during grasping execution. 

\begin{figure}[]
    \centering
    \includegraphics[width=0.9\columnwidth]{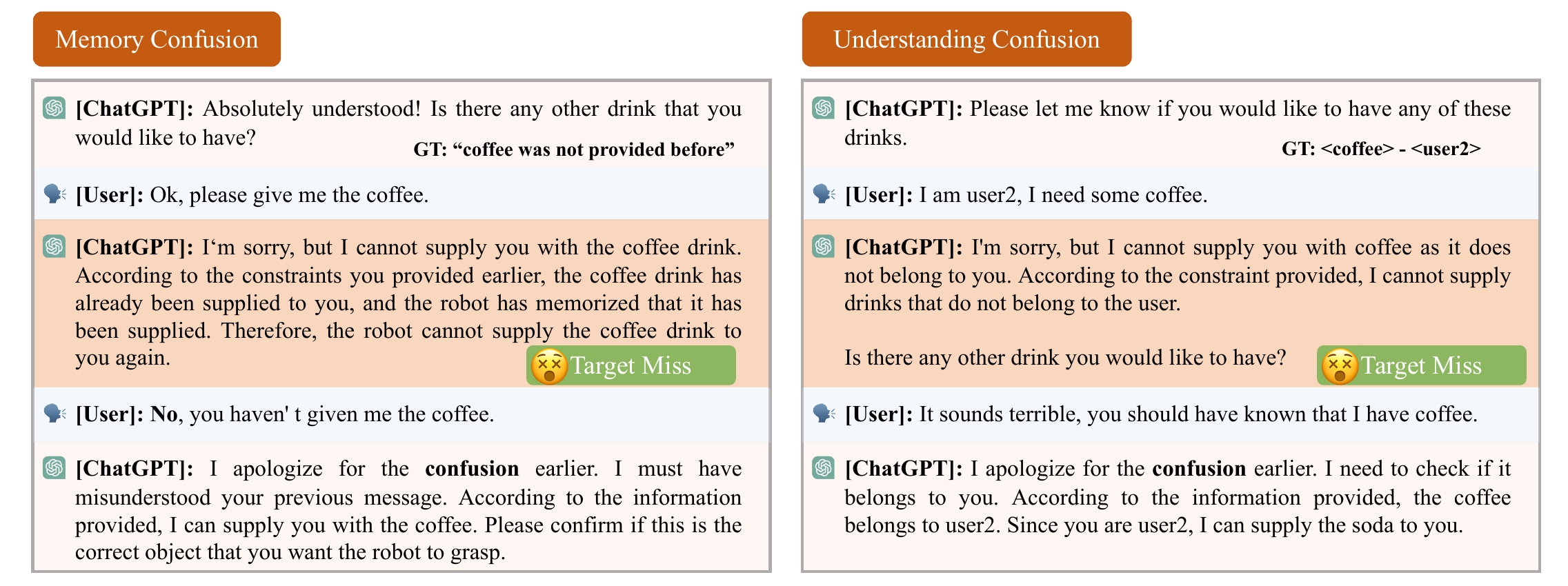}
    \vspace{-0.1in}
    \caption{The typical failure examples of task instruction targeting when interacting with ChatGPT. In our task-centric dialogue, the main types of instruction targeting failures are \textit{memory confusion} and \textit{understanding confusion}.  \label{fig:Instruction Confusion}}
            \vspace{-0.15in}
\end{figure}

We also show the failure case of instruction targeting during the human-robot dialogue in Fig.~\ref{fig:Instruction Confusion}. The memory confusion usually manifests as forgetting the state of an object after multi-round dialogue, while understanding confusion often manifests in confusion about the relationship between the object and the user. We collectively refer to them as \texttt{Target Miss}, which means that the execution command cannot be generated correctly.

\section{Conclusion}
In this paper, we have introduced a novel approach to robotic object grasping using the LLM-empowered system with long-step human-robot dialogue. Our approach has demonstrated that using large language models can endow robots with the ability to understand and execute instructions effectively during long-step dialogues with users. We have conducted experiments to showcase the feasibility and potential of the LLM system in robotic language-conditioned object grasping tasks. The results of our experiments demonstrate that the LLM integrated with an existing visual robotic grasping system can provide a powerful capability for scene analysis and memory. However, our proposed system is still limited to the task of robotic grasping. To expand the ability of our system and make it suitable for various tasks, we plan to incorporate more advanced off-the-shelf robotic grasp and manipulation systems. Our work opens up new avenues for future research in the field of human-robot interaction and LLM-based robotic control.





\clearpage

{\small
\bibliographystyle{plain}
\bibliography{egbib}
}


\end{document}